# Multi-output Classification for Compound Fault Diagnosis in Motor under Partially Labeled Target Domain

Wonjun Yi*[†], Yong-Hwa Park**
*Dept. of Electrical Eng., KAIST., ** Dept. of Mechanical Eng., KAIST.

**Abstract**

This study presents a novel multi-output classification (MOC) framework designed for domain adaptation in fault diagnosis, addressing challenges posed by partially labeled (PL) target domain dataset and coexisting faults in rotating machinery. Unlike conventional multi-class classification (MCC) approaches, the MOC framework independently classifies the severity of each fault, enhancing diagnostic accuracy. By integrating multi-kernel maximum mean discrepancy loss (MKMMD) and entropy minimization loss (EM), the proposed method improves feature transferability between source and target domains, while frequency layer normalization (FLN) effectively handles stationary vibration signals by leveraging mechanical characteristics. Experimental evaluations across six domain adaptation cases, encompassing partially labeled (PL) scenarios, demonstrate the superior performance of the MOC approach over baseline methods in terms of macro F1 score.

## 1. Introduction

Compound fault diagnosis in rotating machinery is essential for operational reliability, but domain shifts and limited labeled data hinder model performance. Conventional multi-class classification (MCC) methods struggle with coexisting faults and label scarcity, reducing adaptability in unsupervised domain adaptation (UDA) scenarios.

This study introduces a multi-output classification (MOC) framework designed to handle partially labeled (PL) target domain dataset by assigning each task-specific layer (TSL) to a specific fault, reducing inter-class interference, improving feature alignment, and compensating data imbalance. The proposed MOC model is inspired by the Tasks-Constrained Deep Convolutional Network (TCDCN) [1], which effectively utilizes TSLs for multi-task learning.

To further improve robustness, we propose frequency layer normalization (FLN), which preserves frequency-domain features in vibration signals. Unlike batch normalization (BN) [2] and layer normalization (LN) [3], FLN effectively minimizes domain shifts caused by varying operating conditions.

Experiments were conducted on a motor based on [4]. The compound fault dataset includes inner race fault (IRF), outer race fault (ORF), misalignment, and unbalance. Since three subsets of datasets are defined by operating condition such as rpm pattern and torque load, six domain adaptation cases were evaluated using macro F1 score. We showed effectiveness of MOC framework and FLN for fault diagnosis under PL conditioned target domain dataset.

## 2. Theory and Experiment

### 2.1 Multi-Output Classification

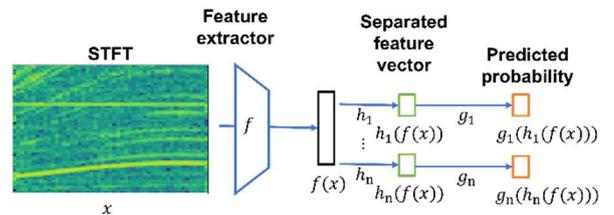

**Fig. 1** Architecture of multi-output classification

We propose the MOC architecture based on TCDCN [1], as shown in Fig. 1. In MCC, a compound fault severity level combination (e.g., IRF 0.2 mm, ORF 0.2 mm, misalignment 0.15 mm, unbalance 10.034 g) is mapped to a single class for classification. In contrast, MOC employs task-specific layers (TSLs), where $h$ and $g$ in Fig. 1 classify the severity levels of feature vector $f(x)$ separately for each fault type. In TSL, $h$ performs feature alignment using UDA loss, such as multi-kernel maximum mean discrepancy (MKMMD) [5], while $g$ functions as a SoftMax classifier, optimizing categorical cross-entropy (CCE) and entropy minimization (EM) loss [6] for classification.

### 2.2 Frequency Layer Normalization

We additionally propose FLN, which is well-suited for the dominance of frequency components at rpm in the Short-Time Fourier Transform (STFT) of the motor, as shown in Fig. 1. Unlike conventional LN [3], which computes the mean and standard deviation across all dimensions for a given sample, FLN performs normalization by calculating these statistics along the frequency dimension.

### 2.3 Experimental Setup

We conducted experiments using the motor setup in Fig. 2, where IRF and ORF occurred in the bearing, while misalignment and unbalance were introduced in the rotor. Severity levels were 0 mm and 0.2 mm for IRF/ORF, 0 mm, 0.15 mm, and 0.3 mm for misalignment, and 0 g, 10.034 g, and 18.070 g for unbalance. Experiments were performed under three conditions: Subset A with a sinusoidal RPM pattern and random torque load, Subset B with a triangular RPM pattern, and Subset C with constant RPM and zero torque load. Each data was sampled at 25.6 kHz in 4 seconds. Fault condition data accounted for only 10% of the normal condition data. In the domain adaptation setup, the source domain dataset was fully composed of labeled data, while the target domain dataset was constituted by 10% of labeled data and 90% of unlabeled data.

---

† Wonjun Yi, lasscap@kaist.ac.kr

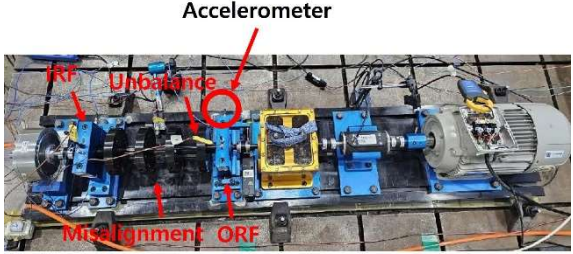

**Fig. 2** Motor used for experiment

During domain adaptation, the model was pre-trained on the source domain for 100 epochs, followed by fine-tuning on the target domain for another 100 epochs. The model parameters with the lowest validation loss in each epoch were selected as the best model.

2.4 Results

As shown in Table 1, MOC achieves a higher macro F1 score across all domain adaptation scenarios compared to conventional MCC.

**Table 1** Macro F1 score based on model architecture

| Source → Target | MCC | MOC |
|---|---|---|
| A → B | 0.991 | 0.999 |
| A → C | 0.673 | 0.774 |
| B → A | 0.827 | 0.807 |
| B → C | 0.833 | 0.870 |
| C → A | 0.766 | 0.797 |
| C → B | 1.000 | 0.996 |
| Average | 0.848 | 0.874 |

Furthermore, the per-fault macro F1 scores indicate that MOC outperforms MCC in classifying all fault types. By assigning classification to separate TSLs for each fault, MOC achieves superior severity level classification across all faults, ultimately enhancing compound fault diagnosis.

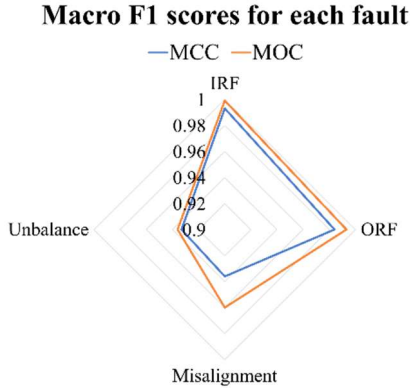

**Fig. 3** Macro F1 scores for each fault
(averaged for all domain adaptation scenarios)

Additionally, as presented in Table 2, FLN demonstrates superior performance in domain adaptation compared to LN and BN within the same MOC framework.

**Table 2** Macro F1 score based on normalization method

| Source → Target | BN [2] | LN [3] | FLN |
|---|---|---|---|
| A → B | 0.984 | 0.805 | 0.999 |
| A → C | 0.810 | 0.623 | 0.774 |
| B → A | 0.661 | 0.588 | 0.807 |
| B → C | 0.782 | 0.699 | 0.870 |
| C → A | 0.705 | 0.574 | 0.797 |
| C → B | 0.960 | 0.864 | 0.996 |
| Average | 0.817 | 0.692 | 0.874 |

3. Conclusion

The proposed MOC framework demonstrated superior performance in domain adaptation, effectively handling compound faults and PL data. Additionally, FLN outperformed BN and LN by leveraging frequency-domain features, improving adaptation to varying operating conditions. Future work may explore advanced model architectures to further enhance MOC performance.

Acknowledgement

This work was supported by the National Research Foundation of Korea (NRF) grant funded by the Korea government (MSIT) (No. RS-2024-00350917).